\documentclass[conference]{IEEEtran}

\IEEEoverridecommandlockouts

\ifCLASSINFOpdf
\else
\fi

\usepackage{graphicx}
\usepackage{cite}
\usepackage[cmex10]{amsmath}
\usepackage{amsmath}
\usepackage{algorithmic}
\usepackage{array}
\usepackage{mdwmath}
\usepackage{mdwtab}
\usepackage{eqparbox}
\usepackage[tight,footnotesize]{subfigure}
\usepackage{stfloats}
\usepackage{url}
\usepackage{booktabs}
\usepackage{import}
\usepackage{color}

\newcommand{\apexstate}    {y_a}
\newcommand{\aas}          {\hat{f}}
\newcommand{\desiredheight}    {y_a^*}

\newcommand{\optAngle}    {\theta_2^*}

\newcommand{\req}[1]          {(\ref{#1})}
\newcommand{\refig}[1]        {Fig.~\ref{#1}}
\newcommand{\resec}[1]        {Section~\ref{#1}}
\newcommand{\retab}[1]        {Table~\ref{#1}}

\DeclareMathOperator*{\argmin}{arg\,min}

\IEEEoverridecommandlockouts
\IEEEpubid{\makebox[\columnwidth]{978-1-4673-7509-2/15/\$31.00 \copyright2015 IEEE\hfill} \hspace{\columnsep}\makebox[\columnwidth]{ }}

\begin{document}
%
\title{Extending The Lossy Spring-Loaded Inverted Pendulum Model with a Slider--Crank Mechanism}


\author{\IEEEauthorblockN{H. Eftun Orhon\textsuperscript{*\dag}, Caner Odaba\c{s}\textsuperscript{*\dag}, \.{I}smail Uyan{\i}k\textsuperscript{\dag}, \"{O}mer Morg\"{u}l\textsuperscript{\dag}, Ulu\c{c} Saranl{\i}\textsuperscript{\ddag}}
	\\
	\thanks{\textsuperscript{*}Marked authors contributed equally.}
	\IEEEauthorblockA{\textsuperscript{\dag}Department of Electrical and Electronics Engineering, Bilkent University, 06800 Ankara, Turkey \\
	\textsuperscript{\ddag}Department of Computer Engineering, Middle East Technical University, 06800 Ankara, Turkey}
		
		\{orhon, canero, uyanik, morgul\}@ee.bilkent.edu.tr, saranli@ceng.metu.edu.tr}



\maketitle

\begin{abstract}

Spring Loaded Inverted Pendulum (SLIP) model has a long history in
describing running behavior in animals and humans as well as
has been used as a design basis for robots capable of dynamic
locomotion. Anchoring the SLIP for lossy physical systems resulted
in newer models which are extended versions of original SLIP with
viscous damping in the leg. However, such lossy models require an
additional mechanism for pumping energy to the system to control
the locomotion and to reach a limit--cycle. Some studies solved
this problem by adding an actively controllable torque actuation
at the hip joint and this actuation has been successively used
in many robotic platforms, such as the popular RHex robot.
However, hip torque actuation produces forces on the COM
dominantly at forward direction with respect to ground, making
height control challenging especially at slow speeds.
The situation becomes more severe when the horizontal speed
of the robot reaches zero, i.e. steady hoping without moving
in horizontal direction, and the system reaches to singularity
in which vertical degrees of freedom is completely lost.
To this end, we propose an extension of the lossy SLIP model
with a slider--crank mechanism, SLIP--SCM, that can
generate a stable limit-cycle when the body is constrained
to vertical direction. We propose
an approximate analytical solution to the nonlinear system dynamics of SLIP--SCM model to characterize its behavior during the locomotion.
Finally, we perform a fixed-point stability analysis on SLIP--SCM
model using our approximate analytical solution and show that proposed model exhibits stable behavior in our range of interest.

\end{abstract}

\IEEEpeerreviewmaketitle

\section{Introduction}
\label{sec:introduction}

The advantages and efficiency of legged morphologies over wheeled and tracked ones on rough terrain is a widely accepted hypothesis among the researchers. The main reason behind this claim is that the legged morphologies are capable of choosing the optimum footholds on the rough terrain, while the wheeled (or tracked) platforms face with the worst case scenario most of the times \cite{Raibert86}. Therefore, designing and constructing legged robots that can negotiate different ground profiles received considerable attention among the robotics researchers \cite{saranli_rhex,wooden_malchano_etal.icra2010}.

One of the most remarkable studies in the area of legged locomotion is that simple spring--mass models, such as the Spring-Loaded Inverted Pendulum (SLIP) model \cite{SchwindPhD98}, can capture COM trajectories of different running animals of varying sizes and morphologies as well as legged robot platforms. However, anchoring the SLIP model for lossy physical robot platforms requires some extensions on the original template such as addition of viscous damping in the leg \cite{ankaralianalytical2009,ndpaper,uyanik.tro2015}.

The main challenge associated with lossy models is the necessity of an additional input to compensate for energy losses of the damping element in the leg in order to sustain a limit cycle. Some studies on one-legged robot platforms running in planar direction use hip torque actuation to inject energy to the system \cite{seipel.ctslip,ankarali_saranli.chaos2010}. However, such actuation methods are infeasible when the horizontal speed of the COM is small and can not even satisfy a rhythmic hopping when the locomotion is constrained to the vertical direction. An ad-hoc solution for this problem is to use linear actuators in series with the compliant leg to supply additional energy to the system \cite{secer.icra2014,byl.icra2013}. However, it requires extensive mechanical revisions on the robot platforms and has a non-negligible effect on system dynamics even if the motor is in the idle mode.

Motivated by the problems in the area, we propose an extension to the lossy SLIP model, when the motion is constrained to vertical direction, with a slider--crank mechanism for energy regulation. Fundamentally, a slider--crank mechanism converts rotary motion to translational motion. Its simple structure allows use of this mechanism for various applications, since the first working examples dated back to as early as $ 3^{rd} $ century \cite{rittichrelief2007}. In modern times this mechanism is utilized for various objectives such as micro/nano robotic applications \cite{Kim.ijrr2006,cutkosky.pttrs2009,hoover.BioRob2010} and biomedical engineering \cite{sup.icorr2009,carrozza.aim2001}.

Actually, our goal is to develop an energetically conservative model for the one-legged hopping robot platform built in our laboratory \cite{uyanik.msthesis} and apply it on this robot by physically constructing the slider--crank mechanism. The use of slider--crank mechanism in legged locomotion applications is also studied before \cite{chang,Andrews.Biomimetics2011}. In \cite{chang}, a slider--crank mechanism consisting of an electromagnetic clutch and a passive trigger with elastic stopper is constructed. Hence, required amount of energy by connecting and disconnecting the clutch repetitively is transferred to the system although motor works continuously. On the other hand, \cite{Andrews.Biomimetics2011} implements a bio-inspired control strategy called `active energy removal (AER)' by using a slider--crank mechanism in order to regulate for energy variations due to the terrain variations. Distinctively, we aim to obtain an analytical solution for the system dynamics of the proposed model and investigate the stability properties.

Despite their simple nature, the stance dynamics of the spring--mass models are non-integrable during the stance phase for planar locomotion \cite{Holmes90}. Addition of a slider--crank mechanism brings additional dynamics to the original model and makes the solutions more complex than already they are. Therefore, we propose an approximate analytical solution for the SLIP--SCM model to represent its COM trajectories for a single stride. Approximate analytical solutions have also been frequently used in literature to solve non-integrable legged locomotion trajectories \cite{schwind.jnls00,geyer.jtb05,ndpaper,yu.bionic2012} with some has been verified in experimental robot platforms \cite{uyanik.tro2015}.

In this paper, we are particularly interested in extending the lossy SLIP model with a slider--crank mechanism in order to obtain an energetically conservative model, meaning that it can inject or remove energy energy from the system. \resec{sec:SLIPSCM_section} details the SLIP--SCM model and its dynamics. \resec{sec:aas} explains the the proposed approximate analytical solution as well as the performed simulation studies to assess its prediction performance. Finally, \resec{sec:case_studies} introduces the dead-beat controller designed to regulate system response and our fixed-point stability analysis.


\section{Vertical SLIP Model with Slider--Crank Mechanism (SLIP--SCM)}
\label{sec:SLIPSCM_section}

\subsection{SLIP--SCM Model}
\label{sec:SLIPSCM_model}

\begin{figure}[b!]
\begin{center}

\includegraphics[width=.95\columnwidth]{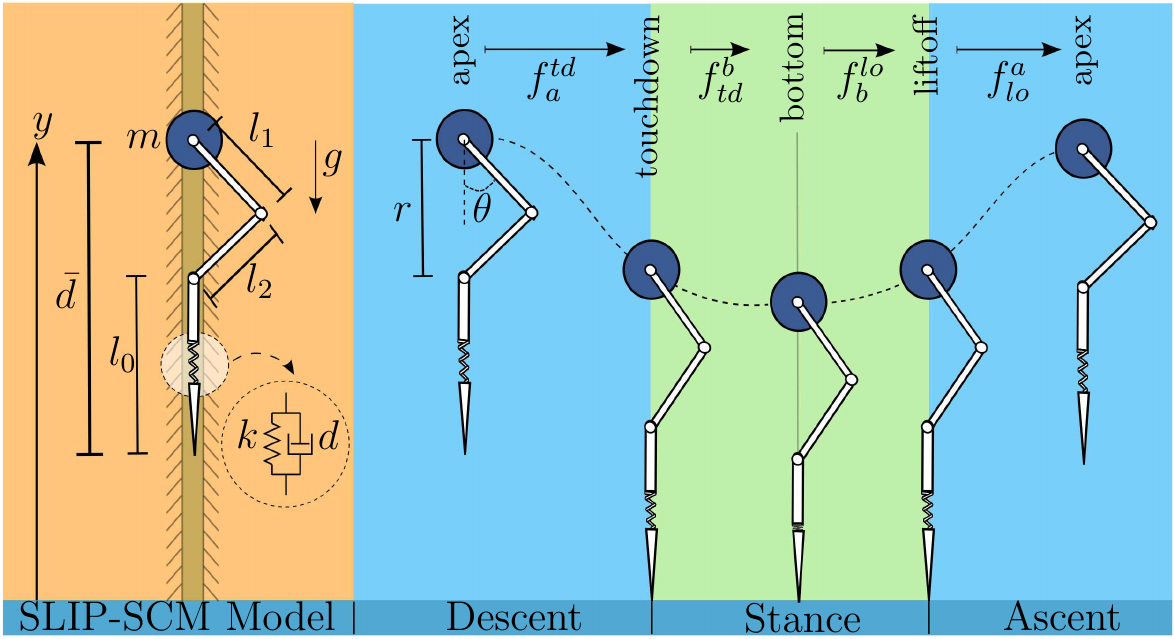}
\caption{\label{fig:system_model}Schematic of the SLIP--SCM	Model ($l_{1}$: crank arm length, $l_{2}$: connecting rod length,  $\bar{d}$: effective damping coefficient, $\theta$: angle between the body and the crank arm)}
\end{center}
\end{figure}

The hopper robot, we consider is simply modeled with a point mass, $m$, attached to a massless compliant leg. The leg consists of a linear spring having a compliance, $k$, and viscous damping, $d$, and connected to robot body through the links of the slider--crank mechanism as illustrated in \refig{fig:system_model}. Notation used throughout the paper is given in \retab{tab:Notation}.

Note that presence of damping in the system results in energy loss, since the initial energy will be exhausted after a certain number of hops. Therefore an additional energy input is required to maintain steady hopping at a desired level. To accomplish this, we use a slider--crank mechanism between the robot body and the leg spring to supply the required energy at each stride by compressing the leg spring to store additional energy in order to compensate damping losses. Similarly, linear actuators are also used for energy regulation purposes in vertical hopping robot models \cite{secer.icra2014, byl.icra2013}.

The SLIP--SCM model has hybrid system dynamics (as the standard SLIP model \cite{SchwindPhD98}); consisting of flight and stance phases of locomotion. The flight phase corresponds to duration when the robot follows a ballistic trajectory during the flight, whereas the stance phase corresponds to duration when the robot is in contact with the ground. The transitions between these phases are determined by touchdown and liftoff events, whose details are illustrated in \refig{fig:system_model}. 

\begin{table}[t!]
	\caption{Notation used throughout the paper}
	\label{tab:Notation}
	\centering
	\setlength\extrarowheight{2.5pt}
	\begin{tabular}{|c|l|} \hline
		\multicolumn{2}{|c|}{\textbf{Extended SLIP-SCM Parameters}}\\
	 	\hline
	 	
		$y$, $\dot{y}$	& Vertical position \& velocity \\
		$k$, $d$ 		& Linear spring compliance \& Viscous damping\\
		$l_0$			& Spring rest length\\
		$m$ 			& Body mass
		\\
		\hline
		\multicolumn{2}{|c|}{\textbf{Return Maps}}\\
		\hline
		$f$				& Numerical solution \\
		$\aas$			& Approximate analytical solution \\
		
		\hline
		\multicolumn{2}{|p{8cm}|}{$\dagger$ \scriptsize{Note that
				subscripts represent the
				system parameters at critical times such as $y_{td}$,
				$y_{b}$, and
				$y_{lo}$ represent the leg position at touchdown, bottom and
				liftoff events, respectively.}}\\ \hline
	\end{tabular}
	\vspace{-3mm}
\end{table}

Being a one-dimensional system, state of SLIP--SCM model can be defined as $Z := \left[ y \;\; \dot{y} \right] $. Then, apex point is defined as the highest point during the flight phase for each stride and associated state for the $k^{th}$ stride is defined as
\begin{equation}
	Z_k := \left[ \apexstate \;\; 0 \right].
\end{equation}
The apex states are critical points for our system, since we discretize the continuous locomotion around these states and design our controllers to regulate system behavior around these points as in most literature studies \cite{ndpaper,ankarali_saranli.chaos2010,uyanik_saranli_morgul.icra2011}.

Although we only control the system response at apex states, we need to characterize the continuous locomotion in order to predict the next apex state. Considering the system dynamics, apex to apex return map of the SLIP--SCM model can be formulated as
\begin{equation}
	\label{eq:complete_map}
	Z_{k+1} = f(Z_k, u) := f_{a}^{td} \circ f_{td}^{b} \circ f_{b}^{lo} \circ f_{lo}^{a}(Z_k),
\end{equation}
\noindent where $f_{a}^{td}$, $f_{td}^{b}$, $f_{b}^{lo}$ and $f_{lo}^{a}$ corresponds to descent, compression, decompression and ascent phases of locomotion, respectively. The control input, $u$, will be applied during the compression phase via changing the crank angle to bring additional compression in the leg spring.

\subsection{System Dynamics}
\label{sec:SLIPSCM_dynamics}

The dynamics of the hybrid SLIP--SCM model is investigated considering the flight and stance phases of locomotion. Assuming the links in the slider--crank mechanism are locked to keep crank angles fixed during the flight phase, system dynamics of the point mass can be written as free fall dynamics under gravity as
\begin{equation}
	\ddot{y} = -g.
\end{equation}

The dynamics of the stance phase is more complex than the flight phase, since both slider--crank and spring--mass dynamics effects the point mass during its locomotion. In order to simplify our analysis, we approach the robot leg as a combination of a compliant leg and a slider--crank mechanism. Therefore, we first assume the links of the slider--crank mechanism are massless as in the standard SLIP model \cite{SchwindPhD98}. Then, we redefine the leg damping as $\bar{d}$ between the point mass and the toe of the leg as illustrated in \refig{fig:system_model} to model energy losses due to slider--crank actuation during the compression and decompression phases. Similar leg damping definitions are also used in literature to support analytical tractability of the dynamic equations \cite{secer.icra2014}.

Based on the aforementioned assumptions, the stance dynamics of the SLIP--SCM model can now be written as
\begin{equation}
	\label{eq:stance_dynamics}
	\ddot{y}=-g-\frac{k}{m}(y-l_{0}-l_{1} \cos(\theta)-l_{2} \cos(\alpha))-\frac{\bar{d}}{m}\dot{y},
\end{equation}
\noindent where $\alpha=\sin^{-1}\left( (l_1 / l_2) \sin \theta \right)$ and represents the angle between the crank arm and the connecting rod.

Note that the crank angle $\theta$ varies during the stance phase, since its angular position is determined by a DC motor rotating slider--crank links. The additional compression starts with an initial crank angle, $\theta_{1}$, and continues to compress the leg spring until it reaches the desired crank angle, $\theta_{2}$, or the bottom event. Therefore, the crank angle for the stance phase can be formulated as
\begin{equation}
	\label{eq:saturated}
	\theta (t) =
	\begin{cases}
		\omega t+\theta_{1}, &\text{if 0\; $\leq$ t \textless $\;t^*$ } \\
		\theta_{2}, &\text{if $t^*$ $\leq$ t $\leq$ $t_{lo}$ }	
	\end{cases}
\end{equation}
\noindent for the cases when the crank angle reaches the desired angle $\theta_{2}$ at time $t^*$. An alternative formulation can be given as
\begin{equation}
	\label{eq:non_saturated}
	\theta (t) =
	\begin{cases}
		\omega t+\theta_{1}, &\text{if 0\; $\leq$ t \textless $\;t_b$ } \\
		\omega t_b+\theta_{1}, &\text{if $t_b$ $\leq$ t $\leq$ $t_{lo}$ }	
	\end{cases}
\end{equation}
\noindent for the cases when compression duration is not sufficient for the DC motor to reach the desired crank angle.

\section{An Approximate Analytic Solution to SLIP--SCM Stance Dynamics}
\label{sec:aas}

In this section, we aim to obtain position and velocity trajectories of the SLIP--SCM model for the stance phase. Therefore, we propose an approximate analytical solution to overcome the issues of nonlinearity in the system dynamics.

\subsection{An Approximate Analytical Solution to Stance Equations}

Our solution for \req{eq:stance_dynamics} begins with a simple assumption that our additional compression on leg is instantaneous. This approximation helps us to simplify our stance equations by removing the additional terms coming from \req{eq:saturated} and \req{eq:non_saturated}. Physically, this corresponds to a DC motor with a high velocity to adjust crank instantly to the desired angle, $\theta_{2}$.

In order to further simplify our equations we define a distance vector, $r$, between the point mass and the leg such that $r_i := l_1 \cos (\theta_i) + l_2 \cos (\alpha_i), \; i = 1,2$. Then, the control action becomes instantaneous change of the distance vector from $r_1$ to $r_2$ at the beginning of the compression phase. The approximated stance equations now take the form
\begin{equation}
	\label{eq:stance_dynamics_app}
	\ddot{y}=-g-\frac{k}{m}(y-l_{0}-r_{2})-\frac{\bar{d}}{m}\dot{y}.
\end{equation}

The approximated form of the stance equations in \req{eq:stance_dynamics_app} resembles an ordinary second order differential equation, whose exact analytic solution is possible. For the present study, we will use the method of undetermined coefficients to solve for position and velocity trajectories during the stance phase.

\begin{figure}[b]
	\begin{center}
		\includegraphics[width=0.95\columnwidth]{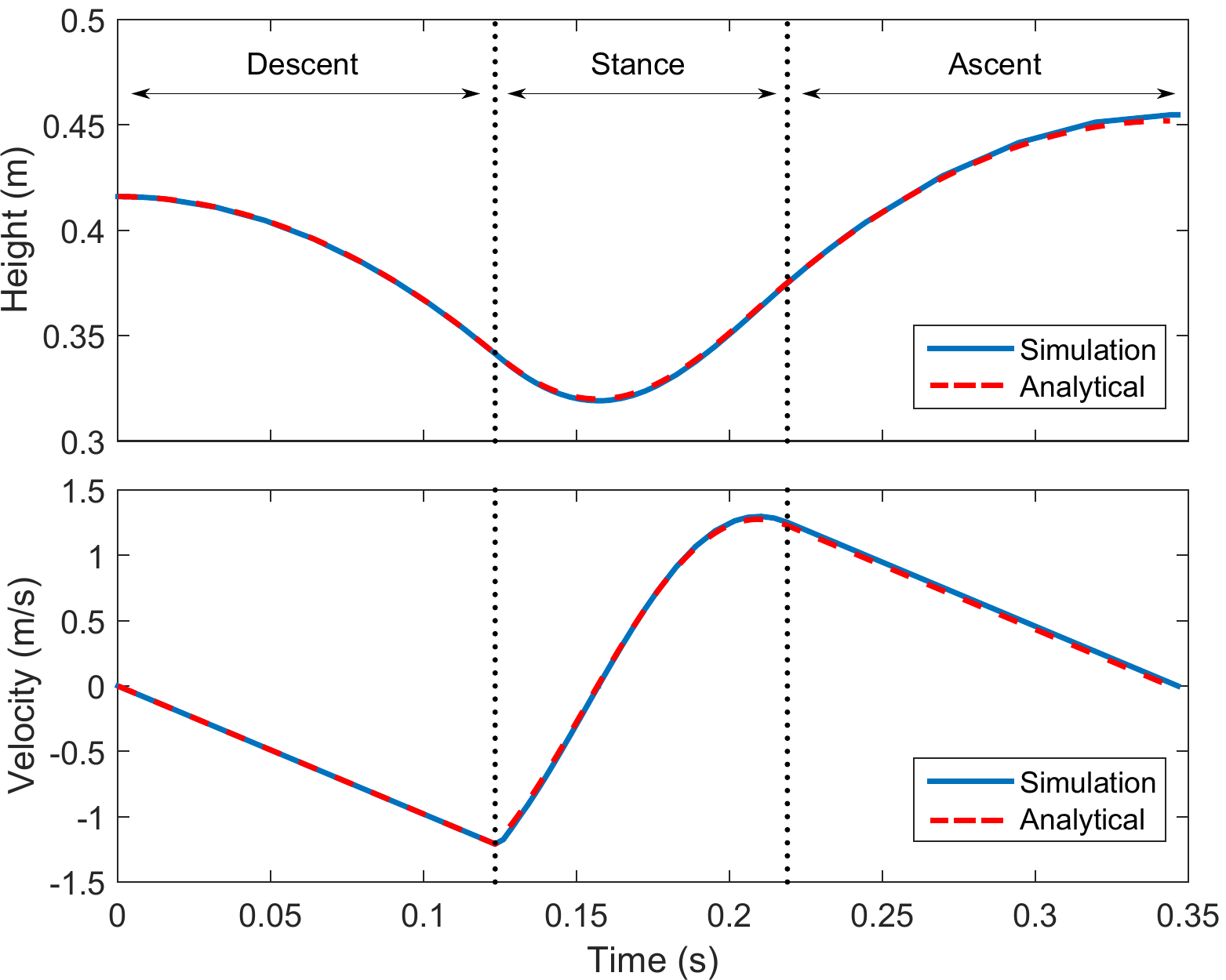}
		\caption{\label{fig:sample_comparison} Comparison of a single stride locomotion between approximate analytical solution and numeric integration of SLIP-SCM dynamics.}
	\end{center}
\end{figure}

In order to ensure liftoff from the ground and enable a hybrid locomotion, the system must be under-damped. Thus, we first assume that $\bar{d}^{2} - 4 m k < 0$ and define the damping ratio, $\xi :=\bar{d} / (2 \sqrt{m k })$, the natural frequency of the system, $w_{0} := \sqrt{ k / m}$, the damped frequency, $w_{d} := w_{0} \sqrt{1- \xi^2}$ and the forcing term, $F := -g + k l_{0} / m + k r_{2} / m$ to solve for \req{eq:stance_dynamics_app}. The solutions for position and velocity trajectories now take form with the new system parameters
\begin{eqnarray}
	\label{eq:pos_solution}
	y (t) &=& e^{-\xi w_{0} t} \left( A_{1} \cos (w_{d} t) + A_{2} \sin (w_{d} t) \right) + F / w_{0}, \quad  \;\\
	\label{eq:vel_solution}
	\dot{y} (t) &=& e^{-\xi w_{0} t} \left( B_{1} \cos (w_{d} t) + B_{2} \sin (w_{d} t) \right).
\end{eqnarray}

\noindent plugging the initial conditions at touchdown state yields the coefficients as
\begin{align}
\notag
 A_{1}&=y_{td}-F/w_{0}^{2}, & A_{2}&=(\dot{y}_{td}+\xi w_{0}A)/w_{d}, \\
\notag B_{1}&=-A_{1}(\xi w_{0}+w_{d}), & B_{2}&= A_{2}(w_{d}-\xi w_{0}).
\end{align}

For a full characterization, we also solve for time of critical events in an approximate analytical nature. \resec{sec:critical_times} details our approximate analytical solution, $\aas$, for touchdown and liftoff times during a single stride. Combining the complete apex to apex return map as in \req{eq:complete_map}, we obtain an analytical predictors for the position and velocity trajectories of the SLIP--SCM model for a single stride. \refig{fig:sample_comparison} illustrates a comparative study where we illustrate the prediction performance of $\aas$ with respect to the numeric solution of the system dynamics in \req{eq:stance_dynamics}.

\subsection{Evaluation of Critical Times}
\label{sec:critical_times}

In this section, we try to find an analytical solution (at least an approximate one) for the critical times on the stance phase which are touchdown time, $t_{td}$, when the hybrid dynamics switches to stance phase, the bottom time, $t_b$, when we stop applying additional compression on the leg spring if it didn't already reach the desired value, and the liftoff time, $t_{lo}$, which determines the end of the stance phase.

Since the model follows a ballistic trajectory during descent phase, the solution for $t_{td}$ is pretty straightforward and can be computed as

\begin{equation}
	t_{td} = \sqrt{2 (y_{0} - r_{1} -l_{0}) / g},
\end{equation}

The second critical time we want to identify is the bottom time, which corresponds to time instant when the maximal leg compression occurs, $\dot{y} = 0$. Fortunately, an analytical solution is available for the bottom time when we use \req{eq:vel_solution} to solve for $t_b$ when $\dot{y}(t_b) = 0$. The solution for $t_b$ can be simply found via trigonometric equalities as
\begin{equation}
	\label{eq:t_b}
	t_{b} =  \tan^{-1} \left(  \frac{A_{2} w_{d} - A_{1} \xi w_{0}} {A_{1} w_{d} + A_{2} \xi w_{0} } \right) /w_{d}.
\end{equation}
Note that solution of \req{eq:t_b} generates infinitely many time values but we use the one that lies between $t_{td}$ and $t_{lo}$.

We complete our discussion by finding an approximate solution for the liftoff time, which can be determined by equating the net force acting on the leg to zero as
\begin{equation}
	h(t) := k(y(t) - l_{0} - r_{2}) + d \dot{y} (t) = 0.
\end{equation}

However an exact solution for $t_{lo}$ is not possible due to the damping in the leg, since presence of damping yields multiple conditions for liftoff event \cite{ndpaper}. Therefore, we utilize a simple approximation strategy to estimate damping time by using famous Newton--Raphson method. Although it is an iterative numerical method, we only use the first iteration as an approximation solution by choosing $t_{lo}^0 = 2 t_b$ as initial guess. Then, solution for liftoff time can be computed as
\begin{equation}
	t_{lo} = t_{lo}^0 - h (t_{lo}^0) \; / \;  \dot{h} (t_{lo}^0).
\end{equation}

\subsection{Assessing Predictive Performance}
\label{sec:predictive_performance}

Although \refig{fig:sample_comparison} yields promising results about the performance of $\aas$, we need to assess its predictive performance for a wide range of initial condition, $y_a$, and control input, $\theta_2$. We choose the initial condition (apex height) for our simulation studies in the range $[l_{rest}, 2 l_{rest}]$, where $l_{rest}$ represents the rest length of the robot when the slider--crank system completely stretches as $l_{rest} = 0.4 \; m$. This choice is based on both biological observations and physical limitations of the robot platforms, since exceeding this threshold results in huge initial energy. For $\theta_2$, we benefit from simulation studies to ensure a stable fixed point for the initial condition range. Our manual calibration tests result in control input in the range $[15^o, 45^o]$. Both the initial condition and control input ranges as well as system parameters used for our simulation studies are given in \retab{tab:SimulationParameters}. The robot parameters are chosen to be consistent with our one-legged hopping robot platform \cite{uyanik.tro2015}.

\begin{table}[ht]
  \vspace{-3mm}
  \centering
  \caption{Initial Condition Ranges and Simulation Parameters}
  \label{tab:SimulationParameters}
  \setlength\extrarowheight{1.5pt}
  \begin{tabular} {|c|c|c|c|c|c|c|}
  \hline
  $y_a$ & $\theta_2$ & k  & $\bar{d}$  & m & $l_0$ & $l_1, l_2$ \\
  ($m$) & ($^o$) & ($N/m$) & ($Ns/m$)  & ($kg$) & ($m$) & ($m$)\\
  \hline \hline
  [0.4, 0.8] & [$15^o$, $45^o$] & 2500 & 10 & 3 & 0.2 & 0.1\\
  \hline
  \end{tabular}
  \vspace{-1mm}
\end{table}

Our goal is to evaluate the prediction performance of $\aas$ with respect to numerical solution, $f$. To accomplish this, we first define two error metrics 
$E_{ap}$ and $E_{lv}$ as 
\begin{equation}
	\label{eq:pos_error}
	E_{ap} := 100 \times \frac{ |y_a - \hat{y}_a |}{y_a}, \quad \, E_{lv} := 100 \times \frac{ | y_{lo} - \hat{y}_{lo} |}{y_{lo}} 	
\end{equation}
\noindent where $E_{ap}$ and $E_{lv}$ correspond to percentage prediction errors for apex position and liftoff velocity, respectively. Note that we compare liftoff velocity, since the vertical velocity of the apex state will be zero by definition. 

\begin{figure}[b!]
	\begin{center}
		\includegraphics[width=0.95\columnwidth]{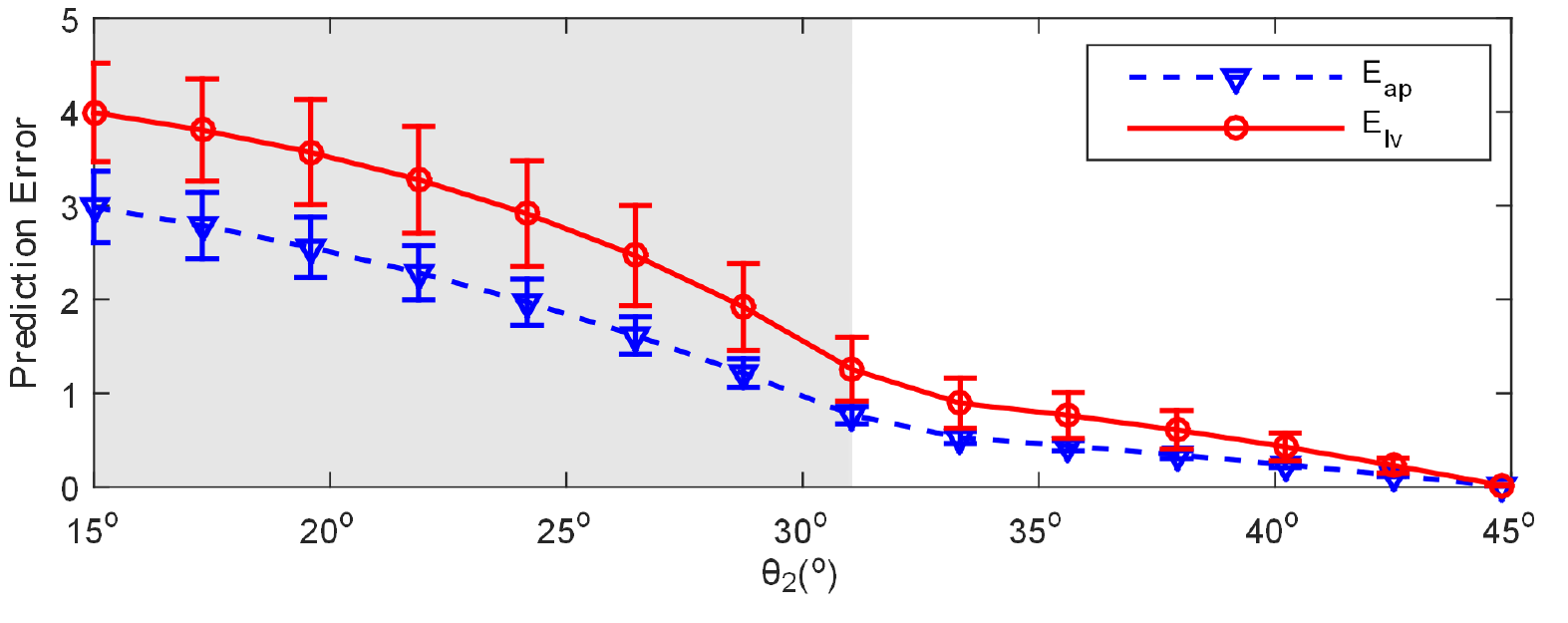}
		\caption{\label{fig:errfig} Percentage position and velocity prediction errors vs. $\theta_2$.}
	\end{center}
\end{figure}

Considering the simulation parameter ranges listed in \retab{tab:SimulationParameters} and error metrics defined in \req{eq:pos_error}, we performed extensive simulation studies to assess prediction performance of $\aas$ by running 10000 unique tests ($100 \times 100$ tests in both ranges). The resulting error across all experiments are computed as 
\begin{equation}
	\label{eq:pos_error_results}
	E_{ap} = 1.31 \pm 1.05, \quad  \, E_{lv}  = 1.93 \pm  1.44
\end{equation}
\noindent Although these errors are relatively high as compared to similar approximate analytical solutions in literature \cite{geyer.jtb05,schwind.jnls00,ndpaper,yu.bionic2012}, they can be easily compensated with adaptive controllers such as the one proposed in \cite{uyanik_saranli_morgul.icra2011}. Additionally, we use deadbeat controllers to regulate system output as will be explained in \resec{sec:deadbeat_control}, which results in very low steady state tracking errors with respect to our prediction errors.

Finally, we project our prediction results onto $\theta_2$ in order to investigate the effect of our control action on the mean prediction error. \refig{fig:errfig} illustrates prediction error vs. the control input with mean and standard deviations as error bars.

Our results show that there is an increasing trend in mean prediction error for both position and velocity variables as $\theta_2$ decreases. This information is useful to design optimal controllers to reduce prediction errors for some applications. More importantly, the behavior in \refig{fig:errfig} can be divided into two classes, since prediction error has a rapid jump when $\theta_2 < 32^o$. This difference occurs based on the sufficiency of DC motor speed to reach desired angular position, $\theta_2$. The white region corresponds to the case when the DC motor reaches $\theta_2$ during the compression phase as in \req{eq:saturated} and the shadowed region represents the part when the DC motor cannot reach $\theta_2$ during the compression as in \req{eq:non_saturated}. Using a better actuator will definitely enlarge the white region in \refig{fig:errfig}, however, it may not be realistic for physical applications. Our goal here is to present a possible problem that is highly likely to be faced when using slider--crank mechanisms for energy injection in legged locomotion models.

\section{Dead-beat Controller and Stability Analysis}
\label{sec:case_studies}


\subsection{Dead-Beat Controller}
\label{sec:deadbeat_control}

\begin{figure}[t!]
 	\begin{center}
 		\includegraphics[width=0.85\columnwidth]{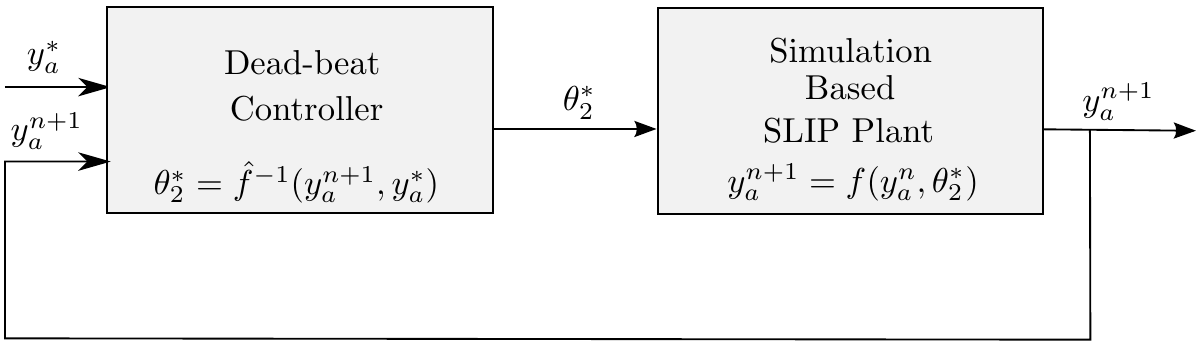}
 		\caption{\label{fig:DeadBeatDiagram} Block diagram representation of the closed-loop system structure. }
 	\end{center}
 \end{figure}

\begin{figure}[b!]
 	\begin{center}
 		\includegraphics[width=0.8\columnwidth]{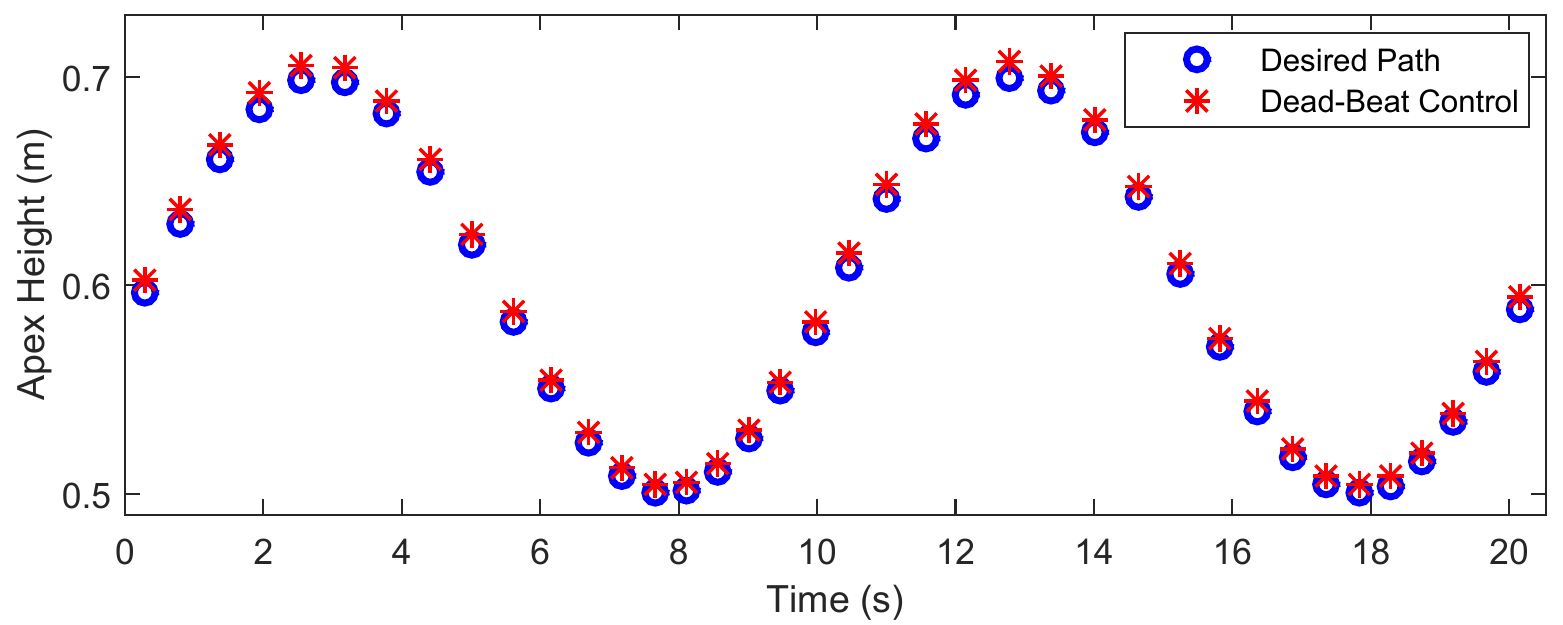}
 		\caption{\label{fig:sine_track} Apex tracking performance of sinusoidal reference trajectory.}
 	\end{center}
 \end{figure}

Our goal in this section is to design a closed-loop controller for the SLIP--SCM model to regulate its output during the locomotion. Therefore, we propose a dead-beat control strategy to control the apex heights of the SLIP--SCM model for each stride. Our motivation here is to utilize $\aas$ instead of numerically solving actual system dynamics in \req{eq:stance_dynamics}. 


\refig{fig:DeadBeatDiagram} illustrates the block diagram of the proposed dead-beat control strategy for SLIP--SCM model. The simulation-based SLIP--SCM plant uses the original, nonlinear system dynamics in \req{eq:stance_dynamics} to simulate locomotion. However, the dead-beat controller block uses $\aas$ to generate control inputs, $\theta_2^*$.

The goal of the controller block is to find optimum control input, $\optAngle$, such that $y_a^*,$ $y_a^{n+1}$ by using $\optAngle=\aas^{-1}(y_a^*,y_a^{n+1})$ as illustrated in \refig{fig:DeadBeatDiagram}. However, $\aas$ does not provide an inverse solution of the system, therefore, the dead-beat controller is implemented in the form of an optimization problem as
\begin{equation}
\label{eq:Opt}
	\optAngle = \argmin_{\theta_2} |\aas(y_a^{n+1},\theta_2) - \desiredheight|.
\end{equation}

In order to assess the performance of the dead-beat controller for different apex states, we perform a sinusoidal path tracking test for SLIP--SCM model. \refig{fig:sine_track} shows both the desired sinusoidal path and the output of the dead-beat controller. Note that system output converges to desired sinusoidal path with a mean percentage prediction error $9.5 \times 10^{-3}$.

\subsection{Stability Analysis}
\label{sec:stability_analysis}
 
 \begin{figure}[b!]
	\begin{center}
		\includegraphics[width=0.88\columnwidth]{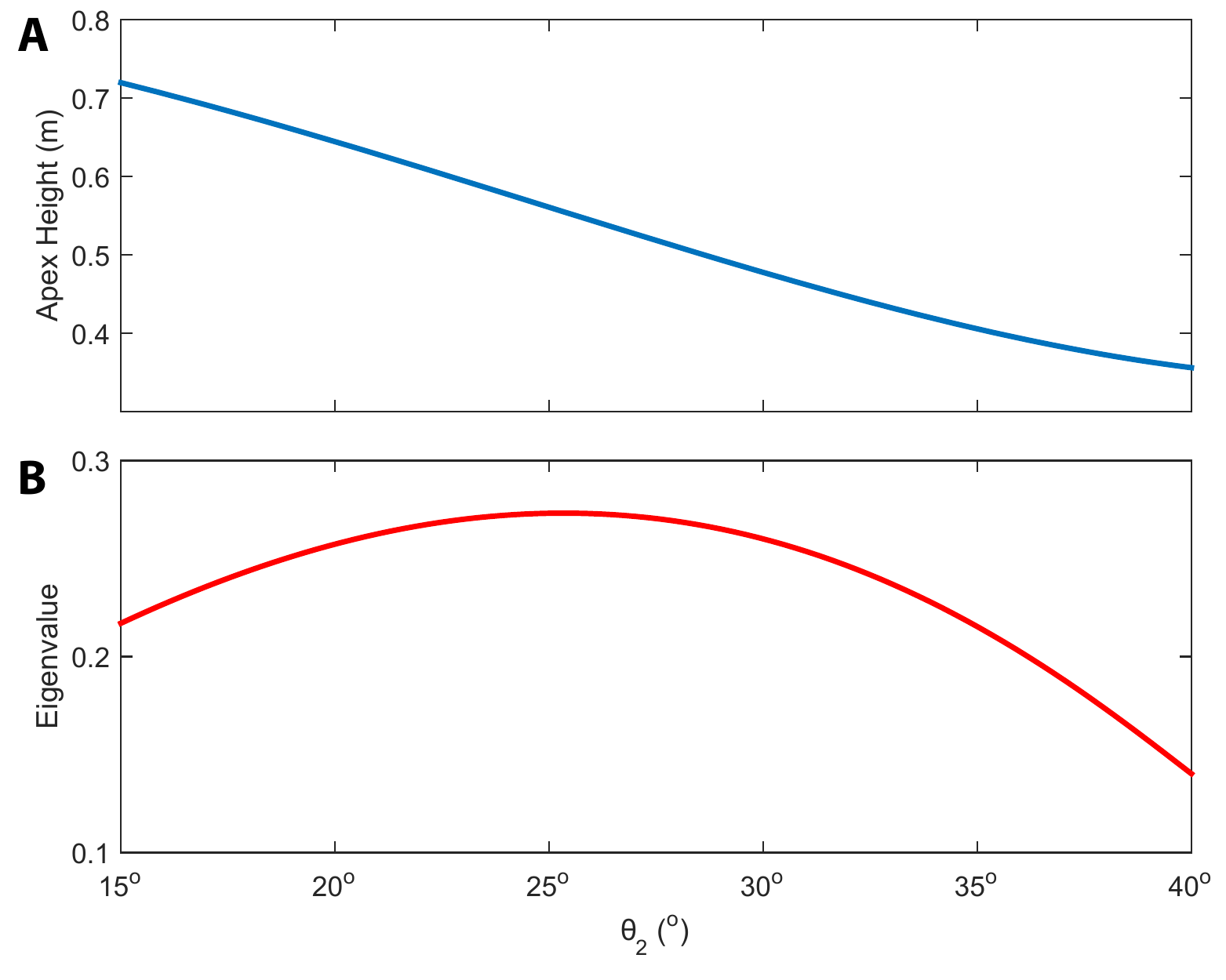}
		\caption{\label{fig:Stability} Fixed-point stability analysis of the SLIP-SCM model with respect to variations on control input.}
	\end{center}
\end{figure}

Although the dead-beat controller explained in \resec{sec:deadbeat_control} allows us to regulate the outputs of the SLIP--SCM model, a fixed-point stability analysis is required for a full characterization of the system. To accomplish this, we investigate fixed-point stability analysis of the system for both the control input, $\theta_2$, and the apex height, $y_a$. Note that both of these stability investigations require to find a fixed point of the system to check system stability around it. Then, stability can be deduced by differentiating $\aas$ with respect to either $\theta_2$ or $y_a$ and checking if the eigenvalues of the Jacobian matrix is inside the unit disk or not. As SLIP--SCM model is a one-dimensional system, we perform a numeric differentiation for our parameter ranges defined in \retab{tab:SimulationParameters} to simplify our derivations.


We first perform fixed-point stability analysis for different control inputs, $\theta_2$ on $\aas$. In order to accomplish this, we first find the initial apex heights yielding fixed-points for each $\theta_2$ in our range of interest as $y_a = \aas (y_a, \theta_2)$.
Note that it is not possible to find fixed-points after a certain degree, since the energy supplied by the controller will not be sufficient to compensate for damping losses and hence will result in a lower apex height. \refig{fig:Stability} (A) shows the apex heights yielding fixed-points for each control input, $\theta_2$. We then perturb $\theta_2$ and find the perturbed response as $\tilde{y}_a = \aas (y_a, \theta_2 + \Delta \theta_2)$. Then the eigenvalue for each control input is computed as
\begin{equation}
	\label{eq:eigenvalue_computation}
	\lambda = \; \mid \tilde{y}_a - y_a  \mid /  \; \Delta \theta_2.
\end{equation}

\noindent \refig{fig:Stability} (B)  shows the eigenvalues of $\aas$ with respect to $\theta_2$. Our system shows stable behavior in our range of interest, since the eigenvalue stays inside the unit disk when we change $\theta_2$.

\begin{figure}[t!]
	\begin{center}
		\includegraphics[width=0.9\columnwidth]{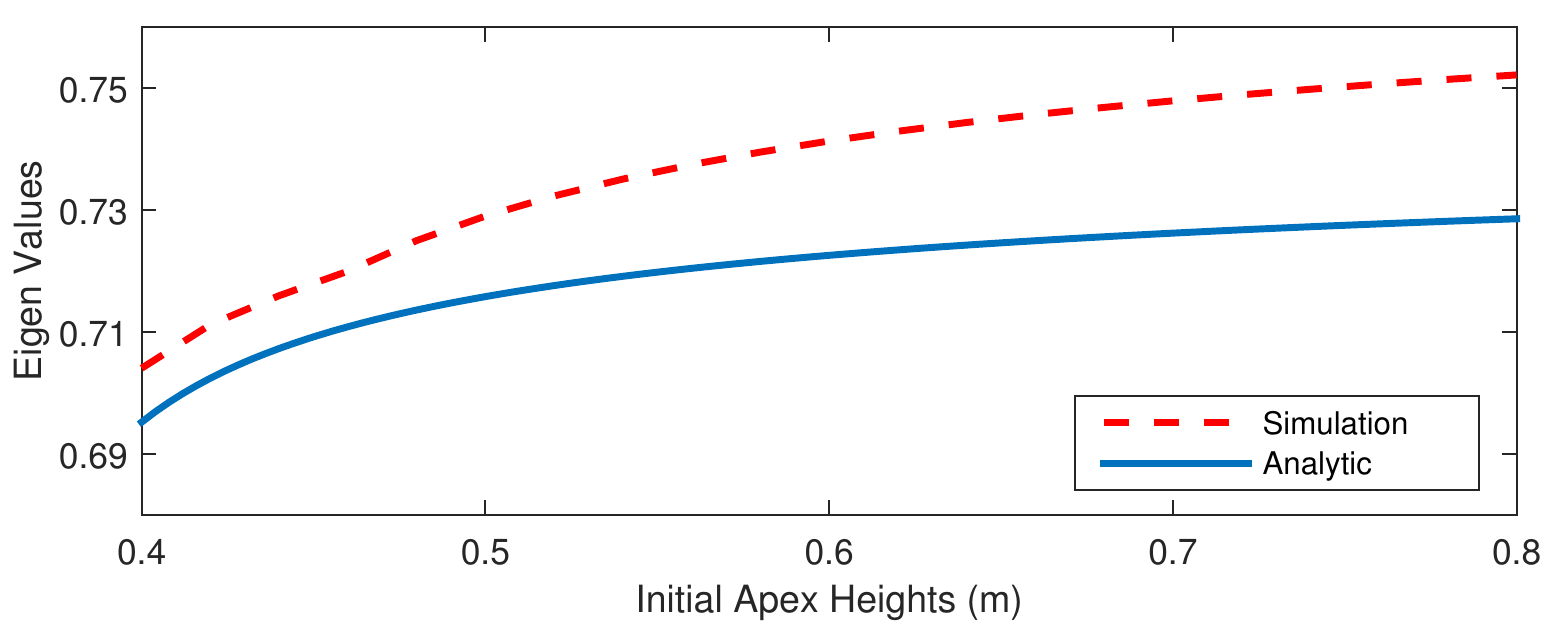}
		\caption{\label{fig:Stability2} Fixed-point stability analysis of the SLIP-SCM model with respect to variations on initial conditions.}
	\end{center}
\end{figure}

Finally, we perform fixed-point stability analysis for different apex heights. Thus, we perturb initial apex height as $\tilde{y}_a = \aas (y_a + \Delta y_a, \theta_2)$ and compute the eigenvalue for fixed control inputs. \refig{fig:Stability2} shows that eigenvalues stay well inside the unit disk (stable response) for our range of interest. We also present a comparison of eigenvalues obtained via both our approximate solution and the numeric integration.

\section{Conclusion}

In this paper, we presented an extension to lossy Spring-Loaded Inverted Pendulum (SLIP) model with a slider--crank mechanism (SLIP--SCM) to obtain an energetically conservative model. The slider--crank mechanism is used to supply additional energy input to our vertically constrained locomotion model during the compression phase by supporting additional compression and hence energy storage on the spring. This additional energy is used to compensate for energy losses due to the damping element in the leg.

Despite their simple nature, the legged locomotion models have non-integrable system dynamics hindering exact analytical solutions to their equations of motion. Addition of slider--crank mechanism to the system aggravates this problem and requires approximations and assumptions on system characteristics to obtain analytical solutions to system dynamics. Thus, we proposed an approximate analytical solution to SLIP--SCM system dynamics and derived the center of mass trajectories of the system for a single stride. Our extensive simulation studies showed that proposed approximate solution is successful in predicting the system response, since the mean prediction error for both position and velocity trajectories stay well below $2\%$.

Finally, we performed a fixed-point stability analysis of the SLIP--SCM model by considering the variations in both the control input and the initial conditions. Our stability analysis showed that SLIP--SCM model, together with our approximate analytical solutions to its dynamics, exhibits a stable behavior in our desired range of interest.
\section{Acknowledgment}
This work is supported by The Scientific and Technological Research Council of Turkey (T\"{U}B\.{I}TAK), through project 114E277. The authors thank ASELSAN Inc. and T\"{U}B\.{I}TAK for \.{I}smail Uyan{\i}k's financial support and Melih \c{C}akmak\c{c}{\i}, G\"{o}rkem Se\c{c}er and Ali Nail \.{I}nal for their ideas and support.

\bibliographystyle{IEEEtran}
\bibliography{references}

\end{document}